\documentclass{ecai} 
\usepackage{latexsym}
\usepackage{amssymb}
\usepackage{amsmath}
\usepackage{amsthm}
\usepackage{booktabs}
\usepackage{enumitem}
\usepackage{graphicx}
\usepackage{color}
\usepackage{algorithmic}
\usepackage{algorithm}

\newcommand{\BibTeX}{B\kern-.05em{\sc i\kern-.025em b}\kern-.08em\TeX}

\newcommand{\intdiv}[2]{\left\lfloor \frac{#1}{#2} \right\rfloor}

\begin{document}

%%%%%%%%%%%%%%%%%%%%%%%%%%%%%%%%%%%%%%%%%%%%%%%%%%%%%%%%%%%%%%%%%%%%%%%%

\begin{frontmatter}

%%% Use this command to specify your submission number.
%%% In doubleblind mode, it will be printed on the first page.

\paperid{6466} 

%%% Use this command to specify the title of your paper.

\title{NITRO-D: Native Integer-only Training of \\ Deep Convolutional Neural Networks}

%%% Use this combinations of commands to specify all authors of your 
%%% paper. Use \fnms{} and \snm{} to indicate everyone's first names 
%%% and surname. This will help the publisher with indexing the 
%%% proceedings. Pplit into "first names" and "surname".
%%% Specifying your ORClease use a reasonable approximation in case your 
%%% name does not neatly sID digital identifier is optional. 
%%% Use the \thanks{} command to indicate one or more corresponding 
%%% authors and their email address(es). If so desired, you can specify
%%% author contributions using the \footnote{} command.

\author[A]{\fnms{Alberto}~\snm{Pirillo}\thanks{Corresponding Author. Email: alberto.pirillo@mail.polimi.it}}
\author[A]{\fnms{Luca}~\snm{Colombo}\thanks{Corresponding Author. Email: luca2.colombo@polimi.it}}
\author[A]{\fnms{Manuel}~\snm{Roveri}} 

\address[A]{Department of Electronics, Information and Bioengineering \\
Politecnico di Milano, Via Ponzio 34/5, Milan, 20133, Italy}

%%% Use this environment to include an abstract of your paper.

\begin{abstract}
Quantization is a pivotal technique for managing the growing computational and memory demands of Deep Neural Networks (DNNs). By reducing the number of bits used to represent weights and activations (typically from 32-bit Floating-Point (FP) to 16-bit or 8-bit integers), quantization reduces memory footprint, energy consumption, and execution time of DNNs. However, most existing methods typically target DNN inference, while training still relies on FP operations, limiting applicability in environments where FP arithmetic is unavailable. To date, only one prior work has addressed integer-only training, and only for Multi-Layer Perceptron (MLP) architectures. This paper introduces NITRO-D, a novel framework for training deep integer-only Convolutional Neural Networks (CNNs) that operate entirely in the integer domain for both training and inference. NITRO-D enables training of integer CNNs without requiring a separate quantization scheme. Specifically, it introduces a novel architecture that integrates multiple \textit{local-loss blocks}, which include the proposed \textit{NITRO-Scaling} layer and \textit{NITRO-ReLU} activation function. The proposed framework also features a novel learning algorithm that employs local error signals and leverages \textit{IntegerSGD}, an optimizer specifically designed for integer computations. NITRO-D is implemented as an open-source Python library. Extensive evaluations on state-of-the-art image recognition datasets demonstrate its effectiveness. For integer-only MLPs, NITRO-D improves test accuracy by up to $+5.96\%$ over the state-of-the-art. It also successfully trains integer-only CNNs, reducing memory requirements and energy consumption by up to $76.14\%$ and $32.42\%$, respectively, compared to the traditional FP backpropagation algorithm.
\end{abstract}

\end{frontmatter}

%%%%%%%%%%%%%%%%%%%%%%%%%%%%%%%%%%%%%%%%%%%%%%%%%%%%%%%%%%%%%%%%%%%%%%%%

\section{Introduction}
\label{sec:introduction}
% Quantization in general
In recent years, quantization of Deep Neural Networks (DNNs) has emerged as a pivotal trend in both the scientific and technological domains~\cite{gholami2022survey}. This technique aims at lowering the computational demand and memory requirements of DNNs by reducing the number of bits to represent weights and activations (e.g., from 32-bit Floating-Point (FP) to 16-bit or 8-bit integers). However, existing quantization frameworks for DNNs retain a dependence on FP operations during training. In this perspective, the aim of this paper is to address the following research question: \textit{Is it possible to train arbitrarily deep Convolutional Neural Networks (CNNs) using only integer values for both training and inference?}

% Distinctions from related works
The literature in this field is extensive but highly fragmented. Existing frameworks for integer-only DNNs can be grouped into three main categories. The first, \textit{quantization-aware inference}, applies quantization only at inference time, while training relies on FP operations to compute gradients~\cite{krishnamoorthi2018quantizing, jacob2018quantization, courbariaux2015binaryconnect, rastegari2016xnor, zhu2016trained, banner2018scalable, lee2022quantune, wei2022qdrop, liu2021post, lin2021fq}. The second, \textit{quantization-aware training}, extends quantization to the training process, further exploiting the advantages of integer arithmetic at the cost of additional complexity in the quantization scheme~\cite{zhou2016dorefa, yang2020training, zeng2024unpack, xiao2024neural, chin2021high, jain2020trained, nia2023training, zhao2020efficient, chen2017fxpnet, wang2022niti, ghaffari2022integer}. Despite these advancements, such approaches still partially rely on FP arithmetic or employ custom fixed-point numeric formats that emulate it. The last category, \textit{native integer-only training}, uses integer-only arithmetic during both inference and training. To date, the only solution in this category targets Multi-Layer Perceptron (MLP) architectures~\cite{song2022pocketnn}.

% The proposed solution
This paper introduces NITRO-D, a native integer-only framework for training deep CNNs exclusively employing integer arithmetic. This approach eliminates the need for any explicit quantization steps or FP operations during training. NITRO-D features a novel network architecture composed of multiple local-loss blocks and introduces novel components, such as the \textit{NITRO-Scaling} layer and the \textit{NITRO-ReLU} activation function, which are specifically designed to prevent integer overflow. In addition, NITRO-D employs a novel learning algorithm adapted from Local Error Signals (LES)~\cite{nokland2019training} to enable integer-only training. Specifically, this algorithm relies on two key innovations: the \textit{IntegerSGD} optimizer and the \textit{NITRO Amplification Factor}, which is used to calibrate the learning rate. To the best of our knowledge, NITRO-D is the first framework in the literature to enable native integer-only training of deep CNNs. Thanks to the simplicity of integer arithmetic~\cite{dally2015high}, NITRO-D offers reduced hardware complexity and lower energy consumption. Integer operations can be executed using smaller bit-widths (such as 8-bit and 16-bit), enabling faster execution times and lower memory requirements compared to traditional 32-bit FP operations. NITRO-D is released as an open-source library to the scientific community\footnote{https://github.com/AI-Tech-Research-Lab/nitro-d}.

% Contributions to the field
The main contributions of this work are:
\begin{enumerate}
    \item The NITRO-D architecture, which enables the construction of arbitrarily deep CNNs operating entirely in the integer domain.
    
    \item The NITRO-D learning algorithm, which enables native integer-only training of the NITRO-D architecture without requiring neither quantization schemes nor FP operations.
\end{enumerate}
Experimental results on multi-class image classification benchmarks show test accuracy improvements of up to $+5.96\%$ for integer-only MLP architectures over the state-of-the-art, and demonstrate the ability to train integer-only CNN architectures. Moreover, the proposed NITRO-D framework reduces memory requirements by up to $76.14\%$ and energy consumption by up to $32.42\%$ compared to the traditional FP Backpropagation (BP) algorithm.

% Reading guide
The remainder of this paper is organized as follows. Section~\ref{sec:related_literature} reviews the related literature. Section~\ref{sec:proposed_solution} introduces the proposed NITRO-D framework, and Section~\ref{sec:experimental_results} presents experimental results to evaluate its effectiveness. Conclusions and future works are finally discussed in Section~\ref{sec:conclusions}.

%%%%%%%%%%%%%%%%%%%%%%%%%%%%%%%%%%%%%%%%%%%%%%%%%%%%%%%%%%%%%%%%%%%%%%%%

\section{Related Literature}
\label{sec:related_literature}
Research on integer-only DNN frameworks can be grouped into three categories. The first involves applying quantization solely during inference, specifically to weights and activations. This is the simplest approach and has been extensively explored in Post-Training Quantization (PTQ)~\cite{krishnamoorthi2018quantizing}, Quantization-Aware Training (QAT)~\cite{jacob2018quantization}, and related methods~\cite{courbariaux2015binaryconnect, rastegari2016xnor, zhu2016trained, banner2018scalable, lee2022quantune, wei2022qdrop, liu2021post, lin2021fq}. However, these techniques rely on FP operations throughout the entire training process.

To address this shortcoming, the second category, known as complete quantization, extends quantization to training, encompassing errors, gradients, and weight updates~\cite{zhou2016dorefa, chen2017fxpnet, yang2020training, zeng2024unpack, wang2022niti, ghaffari2022integer, xiao2024neural, chin2021high, jain2020trained, nia2023training, zhao2020efficient}. While this approach typically results in negligible accuracy degradation, it introduces additional complexity and overhead due to the sophisticated quantization scheme. Moreover, these techniques are unsuitable for true integer-only scenarios, as they either partially rely on FP arithmetic~\cite{zhou2016dorefa, yang2020training, zeng2024unpack, xiao2024neural, chin2021high, jain2020trained, nia2023training, zhao2020efficient}, or employ custom fixed-point formats that emulate FP numbers~\cite{chen2017fxpnet, wang2022niti, ghaffari2022integer}, limiting portability and compatibility.

Finally, the third category, known as native integer-only training, employs exclusively integer arithmetic during both inference and training. This approach fully harnesses the benefits of integer operations and enables DNN training in scenarios that do not support FP arithmetic. To date, this technique has been explored only by PocketNN~\cite{song2022pocketnn}, which successfully implemented native integer-only training for MLP architectures.

In contrast, NITRO-D is, to the best of our knowledge, the first framework to achieve native integer-only training of deep CNNs, entirely eliminating the need for quantization schemes and custom fixed-point numeric formats.

%%%%%%%%%%%%%%%%%%%%%%%%%%%%%%%%%%%%%%%%%%%%%%%%%%%%%%%%%%%%%%%%%%%%%%%%

\section{The Proposed Solution}
\label{sec:proposed_solution}
This section introduces NITRO-D. In particular, Section~\ref{subsec:overview} provides an overview of the framework. Section~\ref{subsec:nitrod_architecture} details the proposed architecture, while in Section~\ref{subsec:nitrod_learning_algorithm} the proposed integer-only learning algorithm is presented.

\subsection{Overview}
\label{subsec:overview}
The proposed NITRO-D framework is designed to enable integer-only training and inference of deep CNNs. It builds on the LES algorithm~\cite{nokland2019training}, a variant of BP~\cite{rumelhart1986learning} that partitions a Neural Network (NN) into local-loss blocks to confine gradient propagation. NITRO-D leverages this structure to prevent integer overflow, a phenomenon that typically arises in the backward pass when training DNNs with integer-based BP. The primary novelty of NITRO-D is its adoption of LES in a fully integer-only setting, where gradients, weights, activations, and updates are confined to the integer domain. 

\begin{figure}[t]
    \centering
    \includegraphics[width=\columnwidth]{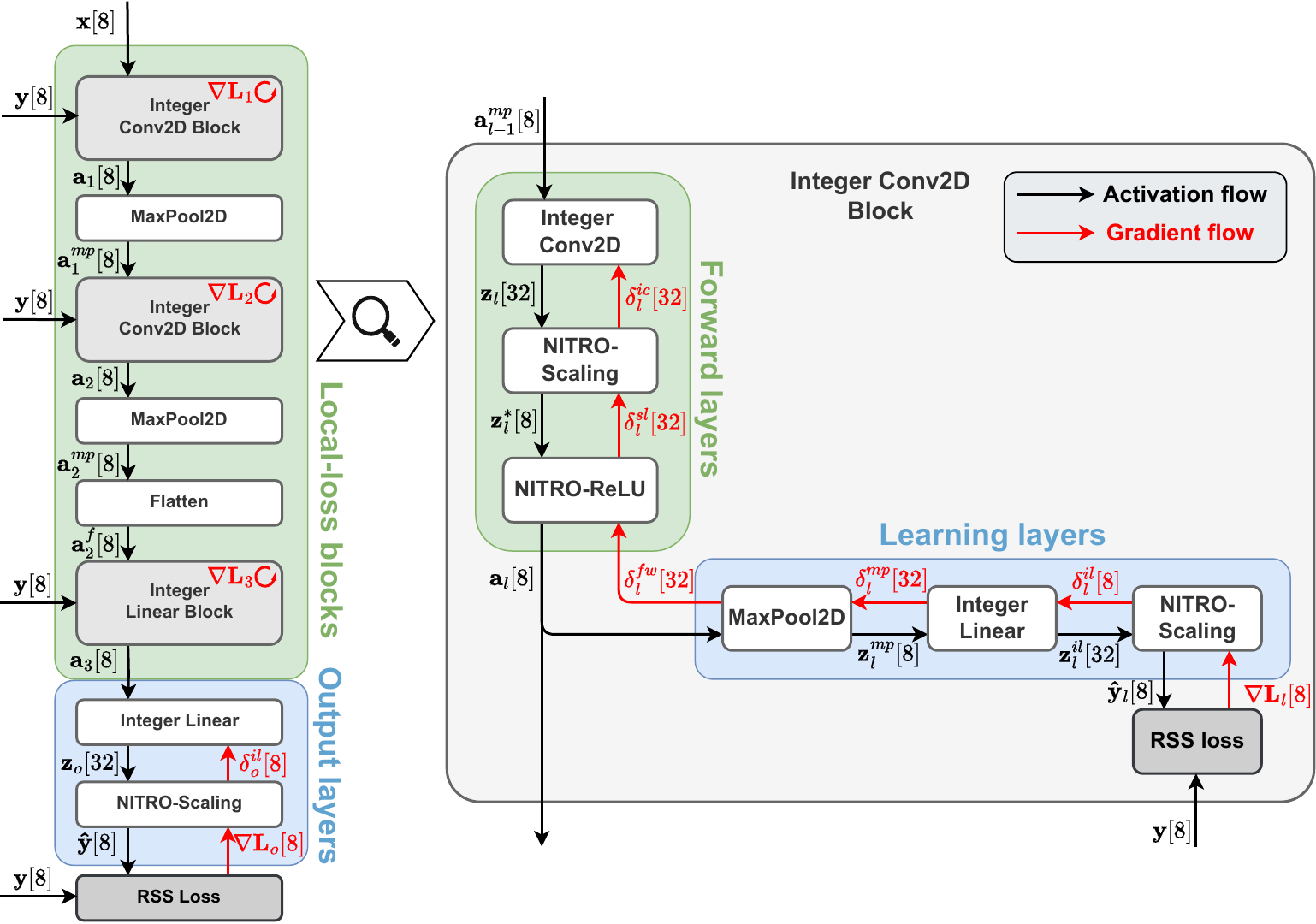}
    \caption{Overview of the NITRO-D architecture for a simple CNN, highlighting the structure of an integer convolutional local-loss block. For each quantity, the required integer bit-width is indicated in brackets.}
    \label{fig:int_local_network}
\end{figure}

The NITRO-D architecture is illustrated in Figure~\ref{fig:int_local_network}, with a focus on its core components: the \textit{integer local-loss blocks}. Each block comprises \textit{forward layers}, which handle the flow of activations during the forward pass, and \textit{learning layers}, which are dedicated solely to training the forward layers using a locally computed loss. By stacking multiple local-loss blocks that can be trained independently, NITRO-D allows for the construction of arbitrarily deep NNs. As shown in Figure~\ref{fig:int_local_network}, the blocks include several components designed to stabilize training by preventing integer overflow and preserving effective propagation of learning signals. More precisely, the \textit{NITRO-Scaling} layer and the \textit{NITRO-ReLU} activation function bound activation magnitudes during the forward pass, while the \textit{NITRO Amplification Factor} maintains consistent weight-update magnitudes across layers. As a result, NITRO-D achieves integer-only training and inference without quantization, entirely eliminating FP operations. This is enabled by \textit{IntegerSGD}, a novel optimizer specifically designed for operating on integer weights and gradients.

\subsection{NITRO-D Architecture}
\label{subsec:nitrod_architecture}
The fundamental component of the NITRO-D architecture is the integer local-loss block. During the forward pass, each block $l = \{1, \dots, L\}$, where $L$ is the total number of blocks, receives the integer activations $\mathbf{a}_{l-1}$ from the previous block $l-1$ and produces new integer activations $\mathbf{a}_l$, which are forwarded to the following block. Concurrently, these activations $\mathbf{a}_l$ are processed by the learning layers, which reduce their dimensionality to match the number of classes of the considered classification task, yielding an intermediate prediction $\mathbf{\hat{y}}_l$. This process repeats across all $L$ blocks until the activations reach the final output layers, where the NN prediction $\mathbf{\hat{y}}$ is produced. Conversely, during the backward pass, gradient flow is confined within each local-loss block $l$. The intermediate prediction $\mathbf{\hat{y}}_l$ and the one-hot ground-truth label $\mathbf{y}$ are used by the local loss function to compute the error $\mathbf{L}_l$ and its gradient $\mathbf{\nabla L}_l$, which are then used to update the parameters of the forward layers in block $l$.

To enable effective integer-only training, NITRO-D introduces two key components: the \textit{NITRO-Scaling} layer and the \textit{NITRO-ReLU} activation function.

\subsubsection{NITRO-Scaling Layer}
\label{par:scaling_layer}
The \textit{NITRO-Scaling} layer is introduced to rescale the integer pre-activations $\mathbf{z}_l$ produced by linear and 2D convolutional layers before they are passed to the \textit{NITRO-ReLU} activation function. This step is necessary to bring the pre-activation values into the operational range of \textit{NITRO-ReLU}, which, as discussed in Section~\ref{par:nitro_activations}, is $[-127, 127]$. Specifically, the \textit{NITRO-Scaling} layer takes the integer pre-activations $\mathbf{z}_l \in \mathbb{Z}$ and produces the rescaled integer pre-activations $\mathbf{z}^*_l \in [-127,127]$, computed as follows:
\begin{equation}
\label{eq:scaling}
   \mathbf{z}_l^{*} = \intdiv{\mathbf{z}_l}{SF_l},
\end{equation}
where $SF_l \in \mathbb{N}$ is the scaling factor.

Selecting an appropriate $SF_l$ is critical to ensure that the pre-activations $\mathbf{z}_l$ are properly scaled. To this end, we analyze how the weights $\mathbf{W}_l$ of layer $l$ transform the input $\mathbf{a}_{l-1}$. For a linear layer, the $i$-th element of the output $\mathbf{z}_l$ is defined as
\begin{equation*}
\mathbf{z}_l^i = \mathbf{a}_{l-1} \mathbf{W}_l^i = \sum_{j=1}^{M_{l-1}} \mathbf{a}_{l-1}^j\ W_l^{ij}
\end{equation*}
where $\mathbf{W}_l^i$ is the $i$-th column of the weight matrix $\mathbf{W}_l$ and $M_{l-1}$ is the dimensionality of the input activations $\mathbf{a}_{l-1}$. For a 2D convolutional layer,
\begin{equation*}
\mathbf{z}_l^i = \mathbf{a}_{l-1} \circledast \mathbf{W}_l^i = \sum_{j=1}^{C_{l-1}}  \sum_{k=1}^{K^2_{l-1}} \ \mathbf{a}_{l-1}^{jk} W_l^{ijk},
\end{equation*}
where $C_{l-1}$ and $K_{l-1}$ are the number of input channels and the kernel size, respectively.

From these expressions, an upper bound $b_{\mathbf{z}_l}$ on the bit-width required to represent $\mathbf{z}_l$ can be derived. More specifically, for a linear layer the upper bound is:
\begin{align}
b_{\mathbf{z}_l} &= 
\sum_{j=1}^{M_{l-1}} \mathcal{O}(b_{\mathbf{a}_{l-1}} + b_{\mathbf{W}_l} - 1) \nonumber \\ 
&= \mathcal{O}(b_{\mathbf{a}_{l-1}} + b_{\mathbf{W}_l} - 1 + \log_2{M_{l-1}}),
\label{eq:bound_fc}
\end{align}
while for 2D a convolutional layer it is: 
\begin{align}
b_{\mathbf{z}_l} &= \sum_{j=1}^{C_{l-1}}  \sum_{k=1}^{K^2_{l-1}}  \mathcal{O}(b_{\mathbf{a}_{l-1}} + b_{\mathbf{W}_l} - 1) \nonumber \\ 
&= \mathcal{O}(b_{\mathbf{a}_{l-1}} + b_{\mathbf{W}_l} - 1 + \log_2{K^2_{l-1} + \log_2{C_{l-1}}}),
\label{eq:bound_conv}
\end{align}
where $b_{\mathbf{a}_{l-1}}$ and $b_{\mathbf{W}_l}$ bound the bit-widths required to represent each element of the input $\mathbf{a}_{l-1}$ and weights matrix $\mathbf{W}_l$, respectively. 

For each block $l = \{2,\dots,L\}$, the input $\mathbf{a}_{l-1}$ is the output of \textit{NITRO-ReLU} of the previous block $l-1$, and thus lies in the interval $[-127,127]$. For this reason, $b_{\mathbf{a}_{l-1}} = 8$. Assuming the weights share the same 8-bit range (i.e., $b_{\mathbf{W}_l} = 8$), the bound for an integer linear layer, defined in Equation~\ref{eq:bound_fc}, simplify to
\begin{equation*}
    b_{\mathbf{z}_l} = \mathcal{O}(15 + \log_2 M_{l-1}).
\end{equation*}
Similarly, the bound for a 2D convolutional layer, defined in Equation~\ref{eq:bound_conv}, simplify to
\begin{equation*}
    b_{\mathbf{z}_l} = \mathcal{O}(15 + \log_2 K^2_{l-1} + \log_2 C_{l-1}).
\end{equation*}
Following this analysis, the scaling factor, used by \textit{NITRO-Scaling} in Equation~\ref{eq:scaling} to scale the pre-activations $\mathbf{z}_l$ to $[-127,127]$, is set to
\begin{align*}
SF_l = \left\{ \; \begin{aligned} 
& 2^8 \times M_{l-1}  && \text{ for linear layers} \\
& 2^8 \times K_{l-1}^2 \times C_{l-1} && \text{ for convolutional layers}
\end{aligned}\right.,
\end{align*}
where $2^8$ denotes the target 8-bit scale of the activation range.

Conversely, the backward pass of the \textit{NITRO-Scaling} layer is implemented using the straight-through estimator (STE)~\cite{bengio2013estimating} due to the non-differentiability of integer division. Hence, for each block $l$, the gradients are passed unchanged through the layer, i.e., $\mathbf{\delta}^{il}_l = \mathbf{\nabla L}_l$ and $\mathbf{\delta}^{ic}_l = \mathbf{\delta}^{sl}_l$, with $\mathbf{\nabla L}_l$ being the gradient of the loss function and $\mathbb{\delta}^{sl}_l$ the gradient propagated by \textit{NITRO-ReLU}, as shown in Figure~\ref{fig:int_local_network}. The use of STE does not affect training stability: the forward pass applies a uniform per-feature scale, which does not change the direction of the activation vector.

\subsubsection{NITRO-ReLU Activation Function}
\label{par:nitro_activations}
The proposed \textit{NITRO-ReLU} is a variant of the LeakyReLU activation function~\citep{xu2015empirical}, specifically designed to operate natively on 8-bit integers. Its definition is a piecewise function consisting of four segments: two inherit the LeakyReLU shape, and two arise from saturating the output to the range $[-127, 127]$. This range is the largest zero-centered interval available in the \texttt{int8} datatype, the smallest hardware-supported integer type. Specifically, \textit{NITRO-ReLU} takes the rescaled pre-activations $\mathbf{z}^*_l \in [-127, 127]$ from the \textit{NITRO-Scaling} layer and produces integer activations $\mathbf{a}_l \in [-127, 127]$ as follows:
\begin{align*}
NITRO\text{-}ReLU(x) = \left\{ \; \begin{aligned} 
& \intdiv{\max  \left\{x, -127 \right\}}{\alpha_{inv}} - \mu_{\texttt{int8}} & x < 0 \\
& \min \left\{x,\ 127 \right\} - \mu_{\texttt{int8}} & x \ge 0
\end{aligned}\right.,
\end{align*}
where $\alpha_{inv} = \intdiv{1}{\alpha} \in \mathbb{N}$, with $\alpha \in \mathbb{R^+}$ being the negative slop, and $\mu_{\texttt{int8}}$ is the function's mean. This mean value is used to center the output $\mathbf{a}_l$ around zero, and is calculated by computing the mean of each segment $i = \{1, 2, 3, 4\}$ defined as:
\begin{align*}
\mu_{\texttt{int8}}^i = \left\{ \; \begin{aligned} 
& \intdiv{-127}{\alpha_{inv}} & x < -127 \\
& \intdiv{-127}{2 \alpha_{inv}} & -127 \le x \le 0 \\
& 63 & 0 < x \le 127 \\
& 127 & x > 127
\end{aligned}\right..
\end{align*}
The final mean $\mu_{\texttt{int8}}$ is the average of these four segment means, and subtracting this pre-computed value centers the function's output around zero. This is crucial for a stable training procedure, as zero-centered activations are known to reduce bias shift and improve convergence~\citep{le1991eigenvalues, lecun2002efficient}. In an integer-only setting, this built-in centering is extremely important because standard techniques like Batch Normalization (BN)~\citep{ioffe2015batch} are infeasible. Implementing BN would require differentiable operations and standard deviation calculations that can easily cause overflow with integer arithmetic. Therefore, by providing built-in and zero-cost normalization, \textit{NITRO-ReLU} is particularly suitable for integer-only scenarios.

\subsection{NITRO-D Learning Algorithm}
\label{subsec:nitrod_learning_algorithm}
The ability of NITRO-D to train integer-only deep CNNs stems from its novel learning algorithm. This section first outlines the overall training process and then details its key components: weight initialization, the loss function, the \textit{IntegerSGD} optimizer, and the \textit{NITRO Amplification Factor}.

After initializing the CNN with an integer adaptation of the Kaiming method~\citep{he2015delving}, training begins with the forward pass, which uses only the blocks' forward layers and the network's output layers, shown in Figure~\ref{fig:int_local_network}. At each iteration, a batch of input data $(\mathbf{X}, \mathbf{y})$ is fed to the first block, producing integer activations $\mathbf{a}_1$. This process repeats sequentially: each block $l$ receives activations $\mathbf{a}_{l-1}$ from the previous block until the final activations $\mathbf{a}_L$ reach the output layers, where the NN prediction $\mathbf{\hat{y}}$ is generated. 

At this point, the backward pass begins, involving the learning layers of each block $l$. First, the global loss gradient $\mathbf{\nabla L}_o$ is computed from the final prediction $\mathbf{\hat{y}}$ and the target $\mathbf{y}$. This gradient is then back-propagated to the output layers, whose parameters are updated using the \textit{IntegerSGD} optimizer. Subsequently, local training occurs independently within each block $l$. The block's integer activations $\mathbf{a}_l$ are downsampled by a pooling layer to produce $\mathbf{z}_l^{mp}$. These are fed to the learning layers to generate a local prediction $\mathbf{\hat{y}}_l$, which, together with the target $\mathbf{y}$, is used to compute the local loss gradient $\mathbf{\nabla L}_l$. This local gradient drives the backward pass for the block. First, it is used to update the parameters of the learning layers using the \textit{IntegerSGD} optimizer. Then, the new gradient $\mathbf{\delta}^{fw}_l$ is computed and back-propagated to the forward layers. Finally, the parameters of the forward layers are updated using \textit{IntegerSGD} along with the \textit{NITRO Amplification Factor}. Crucially, the training of each integer local-loss block is independent, as no gradients flow between them, as shown in Figure~\ref{fig:int_local_network}. This independence allows the backward pass for all blocks to be executed in parallel, significantly enhancing training efficiency. Once all blocks complete their backward pass, training proceeds to the next iteration.

\subsubsection{Weight Initialization}
\label{subsec:weight_init}
NITRO-D employs an integer-based adaptation of the Kaiming initialization method~\cite{he2015delving}, which is commonly used with ReLU-like activation functions. Our approach modifies the standard formulation by approximating all operations to be compatible with integer arithmetic. Specifically, the constant $\sqrt{3}$ is approximated using the integer ratio $\intdiv{1732}{1000}$, while the square root of the layer's input dimensionality $\sqrt{fan_{in}}$ is computed using the integer square root algorithm~\cite{warren2013hacker}. The result is then rescaled by a factor of $128$ to map the initial weights to a suitable integer range. With the proposed methodology, the weights for both forward and learning layers are initialized from a discrete uniform distribution $\mathcal{U} \sim (-b, b)$, where the bound $b$ is given by:
\begin{equation*}
    b = \intdiv{128 \times 1732}{\sqrt{fan_{in}} \times 1000}.
\end{equation*}

\subsubsection{Loss Function}
\label{par:loss_function}
The computational limitations associated with integer arithmetic also extend to the selection of the loss function. NITRO-D utilizes the Residual Sum of Squares (RSS), defined as:
\begin{align*}
& \mathbf{L}_l = \sum_{i=1}^{bs} \frac{1}{2}\left(\mathbf{\hat{y}}^i_l  - \mathbf{y}^i\right)^2,
\end{align*}
where $\mathbf{\hat{y}}_l$ is the output of the learning layers of the $l$-th block, $\mathbf{y}$ is the one-hot encoded ground-truth label, and $bs$ is the mini-batch size. This choice is motivated by the straightforward nature of its gradient, which is computed as:
\begin{align}
\label{eq:loss_gradient}
& \mathbf{\nabla L}_l = \sum_{i=1}^{bs} \mathbf{\hat{y}}^i_l  - \mathbf{y}^i.
\end{align}

\begin{algorithm}[t]
    \caption{Proposed \textit{IntegerSGD} with weight decay}
    \label{alg:integer_sgd}
    \begin{algorithmic}[1]
        \REQUIRE{$\gamma_{inv}$ (inverse learning rate), $\lambda_{inv}$ (inverse weight decay rate), $\mathbf{W}_t$ (model weights), N (training epochs)}
        \STATE{$\eta_{inv} = \gamma_{inv} \lambda_{inv}$}
        \FOR{$t=1$ \TO  N}
        \STATE{$\mathbf{\delta}_t \leftarrow \mathbf{\nabla L}_l(\mathbf{W}_{t-1})$}
        \STATE{$\mathbf{\delta}_t \leftarrow \intdiv{\mathbf{\delta}_t}{\gamma_{inv}}$}
        \IF{$\eta_{inv} \ne 0$}
        \STATE{$\mathbf{\delta}_t \leftarrow \mathbf{\delta}_t + \intdiv{\mathbf{W}_{t-1}}{\eta_{inv}}$}
        \ENDIF
        \STATE{$\mathbf{W}_t \leftarrow \mathbf{W}_{t-1} - \mathbf{\delta}_t$}
        \ENDFOR
        \RETURN{$\mathbf{W}_t$}
    \end{algorithmic}
\end{algorithm} 

\subsubsection{IntegerSGD Optimizer}
\label{par:integer_sgd}
The proposed \textit{IntegerSGD} optimizer, described in Algorithm~\ref{alg:integer_sgd}, adapts traditional SGD for native integer-only training and introduces a custom weight decay procedure. To avoid FP multiplication, the learning rate $\gamma$ and weight decay rate $\lambda$ are redefined as their integer reciprocals, $\gamma_{inv} = \intdiv{1}{\gamma} \in \mathbb{N}$ and $\lambda_{inv} = \intdiv{1}{\lambda} \in \mathbb{N}$. Consequently, update steps involving these rates are performed using integer division. In the integer domain, standard weight decay is challenging as the truncation from integer division can render the decay term negligible. To address this issue, \textit{IntegerSGD} reformulates the update rule to handle the contributions of the gradient $\mathbf{\delta}_t$ and the weight decay term $\intdiv{\mathbf{W}_{t-1}}{\lambda_{inv}}$ separately (see Line 6). Furthermore, to control the interaction between the learning and decay rates, a composite inverse rate $\eta_{inv} = \gamma_{inv} \lambda_{inv}$ is introduced (see Line 1). This approach has a straightforward effect: only weights whose absolute value exceeds the threshold $\eta_{inv}$ are penalized by the decay term. Indeed, for any weight $w^{ij}_l$ of the weight matrix $\mathbf{W}_l$ where $|w^{ij}_l| < \eta_{inv}$, the decay term becomes zero, and no regularization is applied.

\subsubsection{NITRO Amplification Factor}
\label{par:lr_amp}
Within a block, the gradients flowing through the forward layers and the learning layers are characterized by different magnitudes. Specifically, the learning layers receive exactly the loss gradient $\mathbf{\nabla L}_l$, whereas the forward layers receive an amplified gradient 
\begin{equation*}
    \mathbf{\delta}^{fw}_l  = \mathbf{\nabla L}_l\ \mathbf{W}^{il \intercal}_l.
\end{equation*}
due to matrix multiplication with the weights $\mathbf{W}^{il}_l$ in the learning layers. Consequently, a learning rate $\gamma_{inv}^{lr}$ that is suitable for the learning layers proves to be too small for the forward layers, leading to disproportionately large weight updates that can lead to divergence of the training process. 

To address this issue, NITRO-D calibrates the forward-layer learning rate $\gamma_{inv}^{fw}$ using the learning rate from the learning layers $\gamma_{inv}^{lr}$ and a shared \textit{NITRO Amplification Factor} $AF \in \mathbb{N}$. Determining $AF$ requires analyzing the gradient $\mathbf{\delta}^{fw}_l$. Following the same approach introduced in Section~\ref{par:scaling_layer} for the \textit{NITRO-Scaling} layer, an upper bound on the bit-width of the gradient $\mathbf{\delta}^{fw}_l$ is computed as: 
\begin{equation}
\label{eq:bound_grad}
    b_{\mathbf{\delta}_l} = \mathcal{O}(b_{\mathbf{\nabla L}_l} + b_{\mathbf{W}_l} - 1 + \log_2 G),
\end{equation}
where $G$ is the number of classes of the considered classification task and $b_\mathbf{\nabla L_l}$ bounds the bit-width of the loss gradient $\mathbf{\nabla L}_l$. 

NITRO-D uses a custom one-hot scheme in which the target class is encoded as $32$ and the remaining $G-1$ entries are equal to $0$. This design, combined with the RSS loss function, ensures that the gradient of the loss $\mathbf{\nabla L}_l$ fits within 6 bits, resulting in $b_{\mathbf{\nabla L}_l} = 6$. Thus, considering the previous assumptions regarding the magnitude of the weights (i.e., $b_{\mathbf{W}_l} = 8$), the bit-width for the amplified gradient defined in Equation~\ref{eq:bound_fc} simplifies to 
\begin{equation*}
    b_{\mathbf{\delta}_l} = \mathcal{O}(13 + \log_2 G).
\end{equation*}
To rescale such magnitudes to the range $[-127,127]$, the \textit{NITRO Amplification Factor} $AF$ is defined as:
\begin{equation}
    \label{eq:amplification_factor}
    AF = 2^6 \times G.
\end{equation}
Because learning rates are represented by their integer reciprocals, reducing the actual forward layers learning rate by a factor $AF$ corresponds to multiplying its inverse by $AF$:
\begin{equation*}
    \gamma_{\mathrm{inv}}^{fw} \;=\; AF \cdot \gamma_{\mathrm{inv}}^{lr}.
\end{equation*}
This adjusted learning rate is used by the \textit{IntegerSGD} optimizer to compute weight updates for the forward layers.

%%%%%%%%%%%%%%%%%%%%%%%%%%%%%%%%%%%%%%%%%%%%%%%%%%%%%%%%%%%%%%%%%%%%%%%%

\section{Experimental Results}
\label{sec:experimental_results}
This section evaluates the performance of the proposed NITRO-D framework on state-of-the-art image recognition datasets. The goals are twofold: \textit{(i)} to show that NITRO-D outperforms the current state-of-the-art native integer-only training framework for MLPs~\cite{song2022pocketnn}, and \textit{(ii)} to demonstrate that its performance on CNNs is comparable to that of traditional FP training algorithms. In particular, we consider three image classification datasets: MNIST~\citep{deng2012mnist}, FashionMNIST~\citep{xiao2017fashion}, and CIFAR-10~\citep{krizhevsky2009learning}. MNIST and FashionMNIST provide 60,000 training and 10,000 test samples of $28 \times 28$ grayscale images. CIFAR-10 contains 50,000 training and 10,000 test samples of $32 \times 32$ RGB images. While our primary focus is on image recognition tasks, we emphasize that NITRO-D is a general framework applicable to any task suited for MLPs and CNNs. The source code is publicly available to the scientific community.

Experiments were conducted on an Ubuntu 20.04 LTS workstation equipped with 2 Intel Xeon Gold 5318S CPUs, 384 GB of RAM, and an Nvidia A40 GPU. Bayesian hyperparameter optimization was used for each configuration to determine the optimal settings that maximize the test accuracy. Specifically, tuned hyperparameters for NITRO-D included: the learning rate $\gamma_{inv}$, the weight decay rates for forward layers $\eta_{inv}^{fw}$ and for learning layers $\eta_{inv}^{lr}$, the dropout rates for integer 2D convolutional blocks $p_{c}$ and for integer linear blocks $p_l$, and the number of input features for the learning layers $d^{lr}$. The complete set of hyperparameters used for each configuration can be found in Appendix~\ref{sec:appendix_experiment_hyperparams}.

Section~\ref{subsec:mlp_experiments} and Section~\ref{subsec:cnn_experiments} present comparisons between the proposed NITRO-D framework and state-of-the-art baselines for MLP and CNN architectures, respectively. Lastly, an ablation study on hyperparameter selection is detailed in Section~\ref{subsec:ablation}.

\begin{table}[t]
\caption{Overview of the MLP and CNN architectures. The number of filters for Integer 2D Convolution (IntConv2D) and the number of features for Integer Linear (IntLinear) layers are reported between parenthesis. All layers except the last are followed by \textit{NITRO-ReLU}. IntConv2D configuration: kernel size $= 3 \times 3$, stride $= 1$, and padding $= 1$. MaxPool2D configuration: kernel size $= 2 \times 2$, and stride $= 2$.}
  \label{table:architectures}
  \centering
  \begin{tabular}{|c|l|}
    \hline
      Name&\multicolumn{1}{c|}{Architecture}\\
      \hline
      MLP 1& IntLinear (784) $\rightarrow$ IntLinear (100) $\rightarrow$ IntLinear (50) \\ & $\rightarrow$ IntLinear (10)\\
      \hline
   MLP 2& IntLinear (784) $\rightarrow$ IntLinear (200) $\rightarrow$ IntLinear (100) \\ & $\rightarrow$ IntLinear (50) $\rightarrow$ IntLinear (10) \\
    \hline
 MLP 3& IntLinear (784) $\rightarrow$ IntLinear (1024) $\rightarrow$ IntLinear (1024) \\ & $\rightarrow$ IntLinear (1024) $\rightarrow$ IntLinear (10)\\
    \hline
 MLP 4 &IntLinear (1024) $\rightarrow$ IntLinear (3000) $\rightarrow$ IntLinear (3000) \\ & $\rightarrow$ IntLinear (3000) $\rightarrow$ IntLinear (10)\\
    \hline
    VGG8B&IntConv2D (128) $\rightarrow$ IntConv2D (256) $\rightarrow$ MaxPool2D \\ & $\rightarrow$ IntConv2D (256) $\rightarrow$ IntConv2D (512) $\rightarrow$ MaxPool2D \\ & $\rightarrow$ IntConv2D (512) $\rightarrow$ MaxPool2D $\rightarrow$ IntConv2D (512) \\ & $\rightarrow$ MaxPool2D $\rightarrow$ Flatten $\rightarrow$ IntLinear (1024) \\ & $\rightarrow$ IntLinear (10)\\
    \hline
      VGG11B& IntConv2D (128) $\rightarrow$ IntConv2D (128) $\rightarrow$ IntConv2D (128) \\ & $\rightarrow$ IntConv2D (256) $\rightarrow$ MaxPool2D $\rightarrow$ IntConv2D (256) \\ & $\rightarrow$ IntConv2D (512) $\rightarrow$ MaxPool2D $\rightarrow$ IntConv2D (512) \\ & $\rightarrow$ IntConv2D (512) $\rightarrow$ MaxPool2D $\rightarrow$ IntConv2D (512) \\ & $\rightarrow$ MaxPool2D $\rightarrow$ Flatten $\rightarrow$ IntLinear (1024) \\ & $\rightarrow$ IntLinear (10) \\
 \hline
  \end{tabular}
\end{table}

\begin{table*}[t]
\caption{Test accuracy on 10 different runs of MLP architectures.}
\label{table:mlp_results}
\centering
\begin{tabular}{|c|c|c|c|c|c|}
    \cline{3-6}
    \multicolumn{2}{c|}{} & \multicolumn{2}{c|}{Integer-only training} & \multicolumn{2}{c|}{FP32 training} \\
    \hline
    Architecture & Dataset & \textbf{NITRO-D} & PocketNN \cite{song2022pocketnn} & FP LES \cite{nokland2019training} & FP BP \\
    \hline
    MLP 1 & MNIST & $97.36 \pm 0.23$ & $96.98$ & - & $98.00$ \cite{song2022pocketnn} \\
    MLP 2 & FashionMNIST & $88.66 \pm 0.46$ & $87.70$ & - & $89.79$ \cite{song2022pocketnn} \\
    \hline
    MLP 3 & MNIST & $98.28 \pm 0.08$ & - & $99.32$ & $99.25$ \cite{nokland2019training} \\
    MLP 3 & FashionMNIST & $89.13 \pm 0.41$ & - & $91.40$ & $91.63$ \cite{nokland2019training} \\
    MLP 4 & CIFAR-10 & $61.03 \pm 0.60$ & - & $67.70$ & $66.40$ \cite{nokland2019training} \\
    \hline
  \end{tabular}
\end{table*}

\begin{table*}[t]
  \caption{Test accuracy on 10 different runs of CNN architectures.}
  \label{table:cnn_results}
  \centering
  \begin{tabular}{|c|c|c|c|c|}
    \cline{3-5}
    \multicolumn{2}{c|}{} & \multicolumn{1}{c|}{Integer-only training} & \multicolumn{2}{c|}{FP32 training} \\
    \hline
     Architecture&Dataset& \textbf{NITRO-D}      & FP LES \cite{nokland2019training}&FP BP \cite{nokland2019training} \\
    \hline
     VGG8B&MNIST        & $99.45 \pm 0.05$&              $99.60$&$99.74$\\
     VGG8B&FashionMNIST& $93.66 \pm 0.40$&             $94.34$&$95.47$\\
  VGG8B&CIFAR-10& $87.96 \pm 0.39$& $91.60$&$94.01$\\
  VGG11B&CIFAR-10& $87.39 \pm 0.64$& $91.61$&$94.44$\\
    \hline
  \end{tabular}
\end{table*}

\subsection{Evaluating NITRO-D on MLP Architectures}
\label{subsec:mlp_experiments}
The objective of the first set of experiments is to compare the performance of MLP architectures trained with the proposed NITRO-D framework against three baselines: the state-of-the-art integer-only MLP framework~\cite{song2022pocketnn}, FP 32-bit (FP32) LES~\cite{nokland2019training}, and the traditional FP32 BP algorithm. The considered MLP architectures, detailed in Table~\ref{table:architectures}, match those used in the baseline studies~\citep{song2022pocketnn, nokland2019training}.

The first two rows of Table~\ref{table:mlp_results} show that NITRO-D outperforms the state-of-the-art integer-only solution for MLPs with both two and three hidden layers on the MNIST and FashionMNIST datasets. Moreover, its accuracy is comparable to traditional FP32 BP, exhibiting only a minimal accuracy degradation of $-0.64\%$ and $-1.13\%$. This gaps reflect the advantages of the categorical cross-entropy loss over RSS for classification tasks, and of Adam over SGD~\cite{adam2014method}. Similarly, the last three rows of Table~\ref{table:mlp_results}, showing the accuracy for two MLPs composed of three hidden layers, confirm that NITRO-D's results are comparable to those obtained with traditional FP32 learning algorithms across all datasets, with accuracy differences ranging from $-0.97\%$ to $-6.67\%$. 

On the other hand, the proposed solution reduces memory requirements due to the use of smaller bit-widths (such as 8-bit and 16-bit) and leverages the benefits of reduced hardware complexity and lower energy consumption due to the inherent simplicity of integer arithmetic over FP arithmetic. Specifically, Table~\ref{table:costs} summarizes the estimated costs in terms of energy consumption and memory requirements per batch and per step during the training of the considered MLPs. The proposed NITRO-D framework reduces memory requirements by up to $74.37\%$ and energy consumption by up to $32.42\%$ compared to the traditional FP32 BP learning algorithm.

These results validate the effectiveness of the proposed NITRO-D framework for integer-only training of MLPs. It achieves superior accuracy over the state-of-the-art baseline and remains comparable to standard FP32 LES and BP learning algorithms, while substantially reducing resource consumption.

\subsection{Evaluating NITRO-D on CNN Architectures}
\label{subsec:cnn_experiments}
The goal of the second set of experiments is to demonstrate that the proposed NITRO-D framework enables native integer-only training of deep CNNs. Experiments are conducted on the same VGG-like architectures used in~\cite{nokland2019training} and summarized in Table~\ref{table:architectures}. The performance of NITRO-D is compared with FP32 LES and BP baselines. 

As shown in Table~\ref{table:cnn_results}, the use of CNNs, as enabled by NITRO-D, significantly outperforms the state-of-the-art native integer-only training solution for MLPs, yielding considerable test accuracy improvements of up to $+2.47\%$ on MNIST and $+5.96\%$ on FashionMNIST datasets. When compared to traditional FP32 learning algorithms, NITRO-D exhibits small accuracy degradation, ranging from $-0.15\%$ to $-7.05\%$. This gap is an expected trade-off resulting from the use of an integer-compatible RSS loss and SGD optimizer instead of the standard categorical cross-entropy and Adam optimizer~\citep{adam2014method}. This trade-off in accuracy is offset by significant gains in efficiency. As shown in Table~\ref{table:costs}, NITRO-D reduces memory requirements by up to $76.14\%$ and energy consumption by up to $30.45\%$ compared to the traditional FP32 BP algorithm.

\subsection{Ablation Study: Selecting the Hyperparameters}
\label{subsec:ablation}
This section investigates the effects of NITRO-D's key hyperparameters. In particular, the weight decay rates for the forward layers $\eta_{inv}^{fw}$ and learning layers $\eta_{inv}^{lr}$ play a crucial role in mitigating overfitting by constraining weight magnitudes. The decision to utilize two separate decay rates stems from the observation that weights in the forward layers are generally larger than those in the learning layers. Given this disparity, and since the proposed decay mechanism only affects weights whose absolute value exceeds the threshold $\eta_{inv}$, as described in Section~\ref{par:integer_sgd}, a single shared rate would lead to sub-optimal regularization. It would either be excessively strong for the learning layers or negligible for the forward layers. 

\begin{figure*}[t]
    \centering
    \includegraphics[width=0.9\linewidth]
    {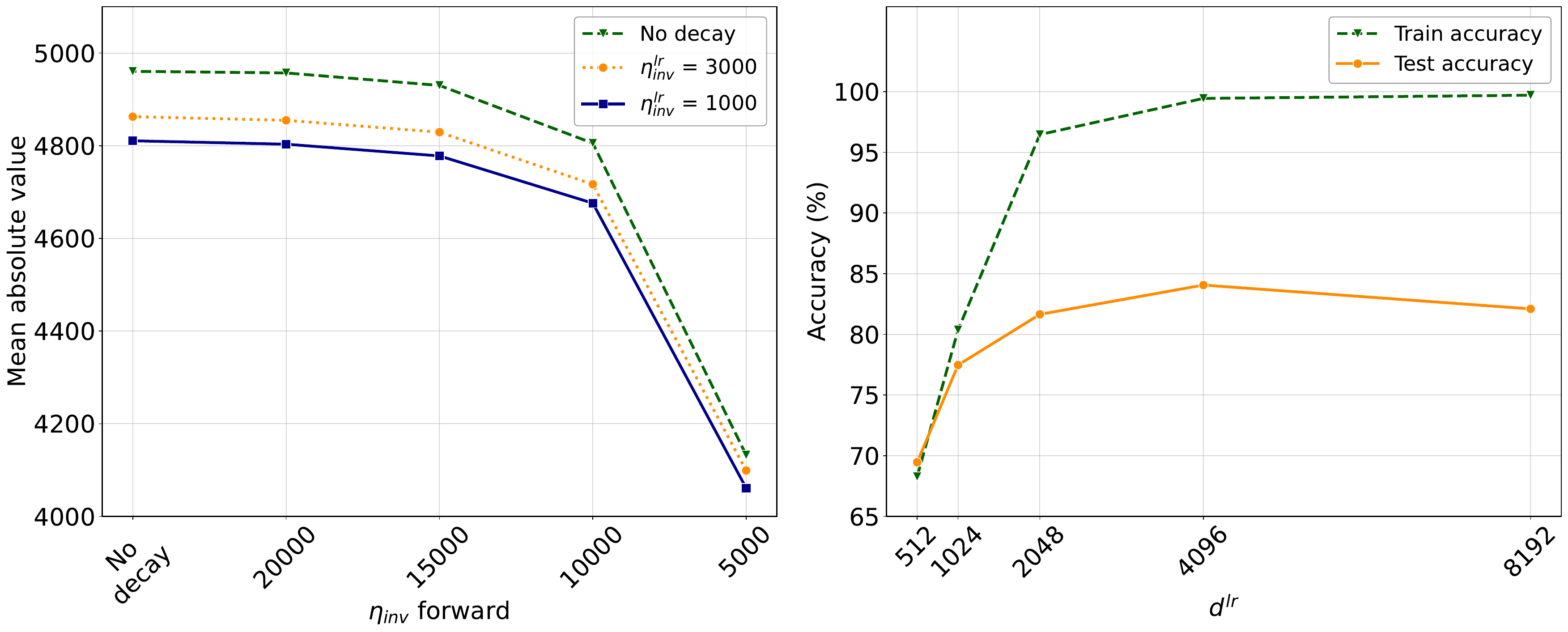}
    \caption{Ablation study of hyperparameters: effect of $\eta_{inv}^{fw}$ and $\eta_{inv}^{lr}$ on weights magnitude (left), and of $d^{lr}$ on accuracy (right). Experiments are conducted with the VGG8B architecture of Table~\ref{table:architectures} on the CIFAR-10 dataset.}
    \label{fig:ablation_hyperparams}
\end{figure*}

Figure~\ref{fig:ablation_hyperparams}-Left shows the impact of different values for $\eta_{inv}^{fw}$ and $\eta_{inv}^{lr}$ on the mean absolute weight value in an integer 2D convolutional layer of the VGG8B model of Table~\ref{table:architectures}. The figure shows that both hyperparameters help to reduce the weights' magnitude. This is because increasing regularization in the learning layers results in smaller gradients being back-propagated to the forward layers. As expected, the highest weight magnitude occurs with no regularization (i.e., ``No decay'' in the figure), whereas the smallest occur when strong regularization is applied to both forward and learning layers.

\begin{table}[t]
  \caption{Estimated training costs for MLPs and CNNs. Memory and energy requirements are computed per batch and per step, accounting for Conv2D and Linear layers operations. FP32 BP of MLP 1 and MLP 2 uses SGD~\cite{song2022pocketnn}, while the other architectures use Adam~\cite{nokland2019training}. Operation energy costs are taken from~\cite{horowitz20141}. $^\dag$: costs computed on the CIFAR-10 dataset.}
  \label{table:costs}
  \centering
  \begin{tabular}{|c|c|c|c|c|}
    \cline{2-5}
    \multicolumn{1}{c|}{} & \multicolumn{2}{c|}{Memory (MB)} & \multicolumn{2}{c|}{Energy ($m$J)} \\
    \hline
    Architecture & \textbf{NITRO-D} & FP32 BP & \textbf{NITRO-D} & FP32 BP \\
    \hline
    MLP 1 & 0.54 & 0.89 & 0.053 & 0.075 \\
    MLP 2 & 1.07 & 1.65 & 0.114 & 0.162 \\
    MLP 3 & 10.28 & 40.11 & 1.813 & 2.675 \\
    MLP 4 & 78.91 & 291.79 & 13.116 & 19.396  \\
    \hline
    VGG8B$^\dag$ & 67.94 & 278.69 & 583.365 & 838.745  \\
    VGG11B$^\dag$ & 92.12 & 386.04 & 861.692 & 1238.883  \\
    \hline
  \end{tabular}
\end{table}

The number of input features for the learning layers $d^{lr}$ directly affects the trade-off between underfitting and overfitting. A low value for $d^{lr}$ can cause underfitting, as information is overly condensed. Conversely, a high value for $d^{lr}$ increases the expressive power of the integer local-loss block, raising the risk of overfitting. Figure~\ref{fig:ablation_hyperparams}-Right indicates that $d^{lr} = 4096$ is a practical starting point that balances this trade-off and provides a solid foundation for further tuning.

Lastly, selecting an appropriate inverse learning rate $\gamma_{inv}$ is essential for convergence. Excessively low values of $\gamma_{inv}$ can cause oversized parameter updates, leading to weights that exceed the expected operational range of the \textit{NITRO-Scaling} and \textit{NITRO-ReLU} layers. Conversely, excessively high values can truncate all weight updates to zero, halting the learning process. Table~\ref{table:lr_results} presents results demonstrating this trade-off for the VGG11B model of Table~\ref{table:architectures} on CIFAR-10. For these experiments, no weight decay or dropout are applied, and the number of input features for the learning layers $d^{lr}$ is set to $4096$. The results indicate that a value of $\gamma_{inv} = 512$ yields optimal performance, making a practical starting point for further tuning.

Although NITRO-D introduces some hyperparameters, finetuning is relatively straightforward. The learning rate $\gamma_{inv}$ and the number of input features for the learning layers $d^{lr}$ can be easily optimized by testing a small grid of discrete values. Tuning the decay rates $\eta_{inv}^{fw}$ and $\eta_{inv}^{lr}$ is guided by the heuristic that weights in the forward layers are typically larger and thus require a larger decay rate than those in the learning layers. Finally, we observed that the dropout rates $p_{c}$ and $p_{l}$ have minimal impact on performance, likely due to the strong regularization already provided by weight decay and the use of low-bit-width integer weights.

\begin{table}[t]
  \caption{Training and test accuracy for the VGG11B architecture of Table~\ref{table:architectures} on the CIFAR-10 dataset for different learning rates $\gamma_{inv}$. No weight decay or dropout are used, while $d^{lr}$ is set to $4096$.}
  \label{table:lr_results}
  \centering
  \begin{tabular}{|c|c|c|}
    \hline
      $\gamma_{inv}$ & Training accuracy (\%) & Test accuracy (\%) \\
      \hline
      256 & (unstable) & (unstable) \\
      512 & 88.86 & 84.66 \\
      1024 & 85.95 & 83.10 \\
      2048 & 72.43 & 70.23 \\    
      4096 & (no learning) & (no learning) \\
   \hline
  \end{tabular}
\end{table}

%%%%%%%%%%%%%%%%%%%%%%%%%%%%%%%%%%%%%%%%%%%%%%%%%%%%%%%%%%%%%%%%%%%%%%%%

\section{Conclusions}
\label{sec:conclusions}
In this paper, we introduced NITRO-D, the first framework to enable native integer-only training of deep CNNs. By design, NITRO-D does not require a quantization scheme, completely eliminating FP operations from both training and inference. The experimental results demonstrate that NITRO-D significantly advances the state-of-the-art by enabling the use of CNNs in this domain. Additionally, extensive comparisons with FP training algorithms for DNNs indicate that, despite its ability to consider integer-only operations on training and inference, NITRO-D provides comparable classification accuracies than FP baselines, while significantly lowering both memory demand and energy consumption. 

NITRO-D has significant implications for deploying DNNs in resource-constrained settings where FP arithmetic is unavailable. This is particularly relevant for tiny machine learning applications targeting low-end devices~\citep{sanchez2020tinyml, immonen2022tiny} and for privacy-preserving machine learning based on homomorphic encryption~\citep{wibawa2022bfv, colombo2023training}.

Future work will focus on three main directions. First, we plan to enhance the framework's efficiency by exploring further bit-width reductions to optimize computational time, memory, and energy consumption. Second, to narrow the performance gap with FP solutions, we will explore improved integer-compatible optimizers with adaptive behaviors, refine the surrogate loss function to better approximate standard training objectives, and investigate learning rate schedulers such as cosine annealing~\cite{loshchilov2016sgdr}. Finally, we will broaden validation by applying NITRO-D to a wider range of modern architectures and deploying it in real-world tiny machine learning and privacy-preserving applications to confirm its practical effectiveness.

%%%%%%%%%%%%%%%%%%%%%%%%%%%%%%%%%%%%%%%%%%%%%%%%%%%%%%%%%%%%%%%%%%%%%%%%

%%% Use this environment to include acknowledgements (optional).
%%% This will be omitted in doubleblind mode.

\begin{ack}
This paper is supported by Dhiria s.r.l. and by PNRR-PE-AI FAIR project funded by the NextGeneration EU program.
\end{ack}

%%%%%%%%%%%%%%%%%%%%%%%%%%%%%%%%%%%%%%%%%%%%%%%%%%%%%%%%%%%%%%%%%%%%%%%%

%%% Use this command to include your bibliography file.

\bibliography{main}

\clearpage
\setcounter{page}{1}

\appendix
\section*{Appendix}

\section{Data Pre-Processing}
\label{appendix:data_preprocessing}
NITRO-D proposes a novel integer-only pre-processing technique designed to transform an integer 2D input dataset $\mathbf{X}$, such as images, into an integer dataset $\mathbf{\hat{X}}$ with mean $\mu = 0$ and standard deviation $\sigma = 64$. This technique allows obtaining a Gaussian distribution of the input data around the range of the \texttt{int8} datatype, also aligning the input data to the operational range of the \textit{NITRO-ReLU}. In a Gaussian distribution, approximately 95\% of the values lie within two standard deviations from the mean. Therefore, a Gaussian distribution with $\mu = 0$ and $\sigma = 64$ ensures that approximately $95$\% of the values are contained in the interval $[-127,127]$, matching the range of the \texttt{int8} datatype. Given the challenges of computing the standard deviation under integer arithmetic, NITRO-D uses the Mean Absolute Deviation (MAD) $\omega$ as a measure of dispersion, as it involves only simple operations that can be carried out effectively with integers.

The first step involves centering the data by subtracting the mean. Then, to obtain $\sigma = 64$ in the pre-processed data $\mathbf{\hat{X}}$, it is required to multiply by 64 and then divide by $\sigma$. For Gaussian data, the MAD $\omega$ can be derived as $\omega \approx 0.8 \sigma$. Therefore, in NITRO-D this rescaling is performed by multiplying the data by $\lfloor 64 \times 0.8 \rfloor = 51$ and then dividing by $\omega$. Thus, the proposed normalization technique employs the following operations: 
\begin{align*}
    \mathbf{\mu}_{int} & = \intdiv{\sum_{i=0}^{N} \mathbf{x}_i}{N}, \\
    \mathbf{\omega}_{int} & = \intdiv{\sum_{i=0}^{N} |\mathbf{x}_i - \mathbf{\mu}_{int}|}{N}, \\
    \mathbf{\hat{x}}_i & = \intdiv{(\mathbf{x}_i - \mathbf{\mu}_{int}) \times 51}{\mathbf{\omega}_{int}} ,
\end{align*}
where $\mathbf{x}_i$ represents the $i$-th element of the dataset $\mathbf{X}$ consisting of $N$ elements. Conversely, $\mathbf{\mu}_{int}$ and $\mathbf{\omega}_{int}$ are the mean and MAD of the dataset computed with integer-only arithmetic, and $\mathbf{\hat{x}}_i$ is the normalized version of the $i$-th element.

Regarding the encoding of the target variable $\mathbf{y}$, NITRO-D relies on a custom version of one-hot encoding, which involves creating a matrix of zeros and then setting to $32$ the elements corresponding to the correct class. This modification allows for greater flexibility in the values that the gradient of the loss $\mathbf{\nabla L}_l$ can take, given that there are no integer values between $0$ and $1$ (and thus, a traditional one-hot encoding would have significantly constrained the gradient's magnitude).

\section{Experiments Hyperparameters}
\label{sec:appendix_experiment_hyperparams}
This section details the full set of hyperparameters utilized to obtain the results presented in Section \ref{sec:experimental_results}, summarized in Table \ref{table:hyperparams}. A batch size of 64 was used across all setups, and training was conducted for 150 epochs. Lastly, the learning rate was reduced by a factor of three whenever the test accuracy reached a plateau.

\begin{table}[H]
\caption{Hyperparameters used to obtain the results reported in Table \ref{table:mlp_results} and Table \ref{table:cnn_results}. $\lambda_{inv}^{fw}$: weight decay rate of \textit{forward layers}, $\lambda_{inv}^{lr}$: weight decay rate of \textit{learning layers}, $d^{lr}$: number of input features of the \textit{learning layers}, $p_l$: dropout rate of Integer Linear Blocks. The learning rate is set to $\gamma_{inv} = 512$, and the dropout rate of Integer Conv2D Blocks is set to $p_c = 0.0$.}
  \label{table:hyperparams}
  \centering
  \begin{tabular}{|c|c|c|c|c|c|c|}
    \hline
     Architecture&Dataset& $\lambda_{inv}^{fw}$&$\lambda_{inv}^{lr}$ & $d^{lr}$& $p_l$\\
     \hline
     MLP 1&MNIST        &  12000&              3000&-&0.00\\
     MLP 2&FashionMNIST& 10000&             8000&-&0.00\\
     \hline
  MLP 3&MNIST& 28000& 5000&-&0.00\\
MLP 3&FashionMNIST& 29000& 6000&-&0.00\\
MLP 4&CIFAR-10     & 19000&           7500&-&0.10\\
    \hline
     VGG8B&MNIST        &              30000&3000& 4096&0.00\\
     VGG8B&FashionMNIST&             28000&3500& 4096&0.10\\
  VGG8B&CIFAR-10& 25000&3000& 4096&0.15\\
  VGG11B&CIFAR-10& 28000&4500& 4096&0.00\\
  \hline
  \end{tabular}
\end{table}

\end{document}